\documentclass[conference]{IEEEtran}
\IEEEoverridecommandlockouts
\usepackage{cite}
\usepackage{amsmath,amssymb,amsfonts}
\usepackage{algorithmic}
\usepackage{graphicx}
\usepackage{textcomp}
\usepackage{xcolor}
\def\BibTeX{{\rm B\kern-.05em{\sc i\kern-.025em b}\kern-.08em
    T\kern-.1667em\lower.7ex\hbox{E}\kern-.125emX}}
\begin{document}

%\title{TrainableTIMtrace: AI-based generation of TIM dispense paths without labeled training data
%\thanks{* A. Steimer and R. Mikut contributed equally to this work.}	
%}
%\author{
%	\IEEEauthorblockN{Simon Baeuerle\textsuperscript{1,2}, Ian F. Mendonca\textsuperscript{2}, Kristof Van Laerhoven\textsuperscript{3}, Andreas Steimer\textsuperscript{4,*}, Ralf Mikut\textsuperscript{1,*}}
%	\IEEEauthorblockA{\textsuperscript{1}Institute for Automation and Applied Informatics (IAI), Karlsruhe Institute of Technology (KIT), D-76344 Eggenstein-Leopoldshafen, Germany. (e-mail: simon.baeuerle@kit.edu)}
%	\IEEEauthorblockB{\textsuperscript{2}Robert Bosch GmbH, D-72762 Reutlingen, Germany.}
%	\IEEEauthorblockC{\textsuperscript{3}Department of Electrical Engineering and Computer Science, University of Siegen, D-57076 Siegen, Germany.}
%	\IEEEauthorblockD{\textsuperscript{4}Bosch Center for Artificial Intelligence (BCAI), Robert Bosch GmbH, D-71272 Renningen, Germany.}
%}
	
%\title{Rapid AI-based generation of coverage paths for dispensing applications
%\title{Rapid AI-based generation of dispense paths for Thermal Interface Materials
%\title{Rapid AI-based generation of coverage paths for fluid dispensing without labeled training data
	
%\title{Rapid generation of coverage paths for dispensing applications using deep neural networks
%\title{Rapid generation of dispense paths for Thermal Interface Materials using deep neural networks

\title{Rapid AI-based generation of coverage paths for dispensing applications
\thanks{* A. Steimer and R. Mikut contributed equally to this work.}
\thanks{\textsuperscript{1} Institute for Automation and Applied Informatics (IAI), Karlsruhe	Institute of Technology (KIT), D-76344 Eggenstein-Leopoldshafen, Germany. (e-mail: ralf.mikut@kit.edu)}% <-this % stops a space
\thanks{\textsuperscript{2} Robert Bosch GmbH, D-72762 Reutlingen, Germany.}% <-this % stops a space
\thanks{\textsuperscript{3} Department of Electrical Engineering and Computer Science, University of Siegen, D-57076 Siegen, Germany.}% <-this % stops a space
\thanks{\textsuperscript{4} Bosch Center for Artificial Intelligence (BCAI), Robert Bosch GmbH, D-71272 Renningen, Germany.}% <-this % stops a space
}

\author{
		\IEEEauthorblockN{Simon Baeuerle\textsuperscript{1,2}, Ian F. Mendonca\textsuperscript{2}, Kristof Van Laerhoven\textsuperscript{3}, Ralf Mikut\textsuperscript{1,*}, Andreas Steimer\textsuperscript{4,*}}
	}

%\author{\IEEEauthorblockN{NN\textsuperscript{1,2,3,4,*}}
%}

\maketitle

\begin{abstract}
	Coverage Path Planning of Thermal Interface Materials (TIM) plays a crucial role in the design of power electronics and electronic control units.
	Up to now, this is done manually by experts or by using optimization approaches with a high computational effort.
	We propose a novel AI-based approach to generate dispense paths for TIM and similar dispensing applications.
	It is a drop-in replacement for optimization-based approaches.
	An Artificial Neural Network (ANN) receives the target cooling area as input and directly outputs the dispense path.
	Our proposed setup does not require labels and we show its feasibility on multiple target areas.
	The resulting dispense paths can be directly transferred to automated manufacturing equipment and do not exhibit air entrapments.
	The approach of using an ANN to predict process parameters for a desired target state in real-time could potentially be transferred to other manufacturing processes.
\end{abstract}

\section{Introduction}
Thermal Interface Materials (TIM) are essential for thermal management in power electronics and electronic control units.
They are applied to a target cooling area with an automated CNC machine during automotive series manufacturing along a path.
This dispense path is compressed during the consecutive joining of the heatsink and the TIM spreads over the cooling area.
Ideally, the compressed state matches well to the cooling area and no material flows beyond the cooling area.
Air entrapments significantly inhibit thermal performance and need to be avoided~\cite{gowda_voids_2004}.
The design of dispense paths for related applications such as adhesive dispensing follows very similar boundary conditions.

\section{Related work}

%Baeuerle and Steimer outline the conceptual approach of \textit{TrainableTIMtrace} in a patent~\cite{baeuerle_waerme_2023}.
%Baeuerle et al. outline a more general formulation of this approach -- which is applicable to a wide range of manufacturing processes -- in another patent~\cite{baeuerle_prozess_2023}.
Baeuerle et al.~\cite{baeuerle_timtrace_2025} propose the optimization-based approach \textit{TIMtrace}, which is successfully validated on real-world products.
A drawback is its long computation time.
Baeuerle et al.~\cite{baeuerle_rapid_2024} train an ANN to serve as flow behavior model for TIM.
This ANN is tested in \textit{TIMtrace} for dispense path planning. In parallel to our research activity, Scholtes et al.~\cite{scholtes_convolutional_2025} recently mention a forward-backward model, which is used to design two-dimensional dispensed states for adhesives.
% -- which is basically the same concept as outlined in the above patents~(\cite{baeuerle_waerme_2023, baeuerle_prozess_2023}).
The underlying forward ANN is similar to the flow behavior model ANN of Baeuerle et al.~\cite{baeuerle_rapid_2024}.
We discuss the limitations of the implementation by Scholtes et al. in Section~\ref{sec:discussion}.
The field of amortization learning is conceptually related.
Instead of optimizing target variables for each setting, the target variables are predicted by a Machine Learning model.
Bitzer et al.~\cite{bitzer_amortized_2023} propose an approach to infer the hyperparameters of a Gaussian process for a given dataset.
For a more extensive elaboration of related Coverage Path Planning approaches, we kindly refer to our recent publication \textit{TIMtrace}~\cite{baeuerle_timtrace_2025}.

We introduce a novel AI-based design generation approach, which infers dispense coverage paths for given target areas.
We directly design the dispense path in an end-to-end fashion and include air entrapments in our objective function.
We show the feasibility of our new approach and discuss its benefits and drawbacks.
\section{Method}
\begin{figure*}[!htb]
	\centerline{\includegraphics[width=.85\textwidth]{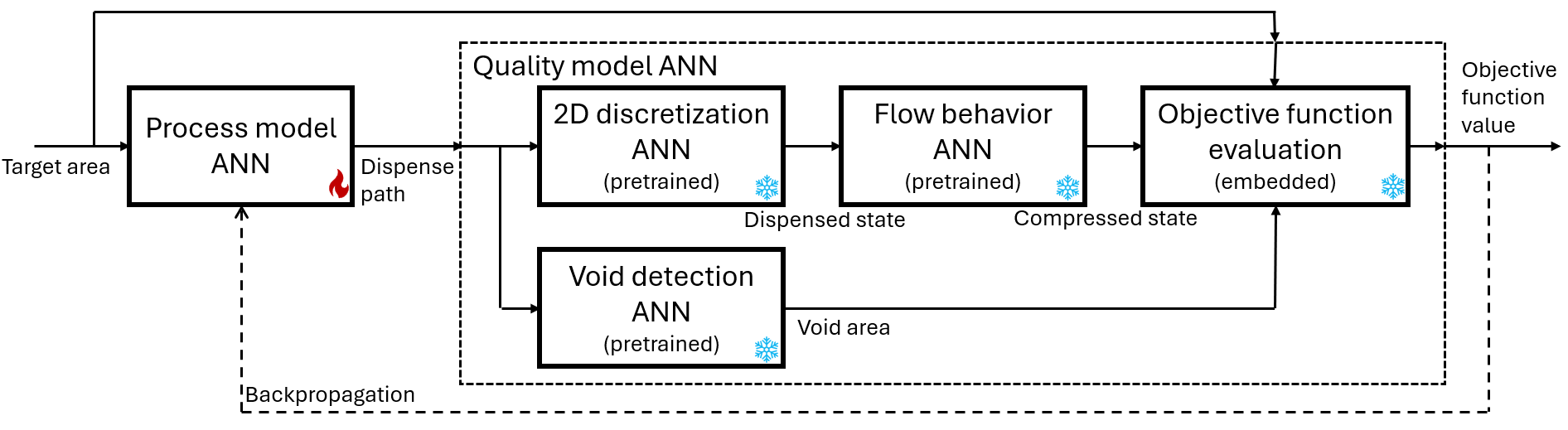}}
	\caption{Approach of our newly proposed training setup. No labeled training data is needed. The process model ANN on the left is trained (indicated by the fire icon) with the help of the quality model ANN on the center-right. The quality model ANN itself is composed of multiple individual models - namely the 2D discretization ANN, the flow behavior ANN and the void detection ANN. During the training of the process model ANN, the weights of the pretrained quality model ANN are frozen (indicated by the ice icon). After training, the process model ANN can predict dispense paths for any given target cooling area.}
	\label{fig:03_approach}
\end{figure*} 

The overall approach with its functional components is shown in Figure~\ref{fig:03_approach}.
The pretrained quality model ANN as shown on the center-right is composed of multiple ANNs.
We utilize the functional components for the 2D discretization, the flow behavior and the void detection from the \textit{TIMtrace} approach to generate the required training data.
%Any other functional component could be appended in a similar fashion, e.g. to integrate further manufacturing constraints.
ANNs are pretrained for each of those functional components and together form a single pretrained quality model ANN.
This pretraining closely follows the procedure as described by Baeuerle et al.~\cite{baeuerle_rapid_2024} for the flow behavior ANN and relies on previously generated training data.

We link a second ANN -- the process model ANN -- to this quality model ANN.
It receives target cooling areas as input and outputs the dispense paths, which are then fed to the quality model ANN.
The output dimensionality of the process model ANN is fixed.
We define 12 outputs, which correspond to the x- and y-coordinates of a polygonal chain with five segments and six path points.
The dispense feedrate $f$ is set to exactly match the theoretically needed TIM amount $V$ for a given target area with $f=V/l$ for a path with length $l$.
The quality model ANN predicts the objective function value for a given dispense path.
This objective function value is used to enable a training of the process model ANN -- even though no training data is available for the dispense paths.

We use geometric outlines of living rooms from floor plans as input target areas to train our model.
These shapes are easily available and similar enough to the actual cooling areas to assess the feasibility of our approach.
The labels are a constant optimum value of zero for the objective function for each input, i.e. our approach does not require specific labeling activities.
We use the following layers to build our process model ANN:
Conv2D (32 3$\times$3 filters), Conv2D (8 3$\times$3 filters), MaxPooling2D, Conv2D (1 5$\times$5 filter), Flatten, Dense (64 neurons), Dense (256 neurons), Dense (12 neurons).
Training is carried out using an \textit{Adam} optimizer with a learning rate of 0.000574 and a batch size of 8.

%We employ an automated hyperparameter optimization during the training procedure for the architecture of the process model ANN.

After training, the process model ANN is disconnected from the quality model ANN and can predict dispense paths for new target areas.

\section{Results}
\begin{figure}[!htb]
	\centerline{\includegraphics[width=.8\columnwidth]{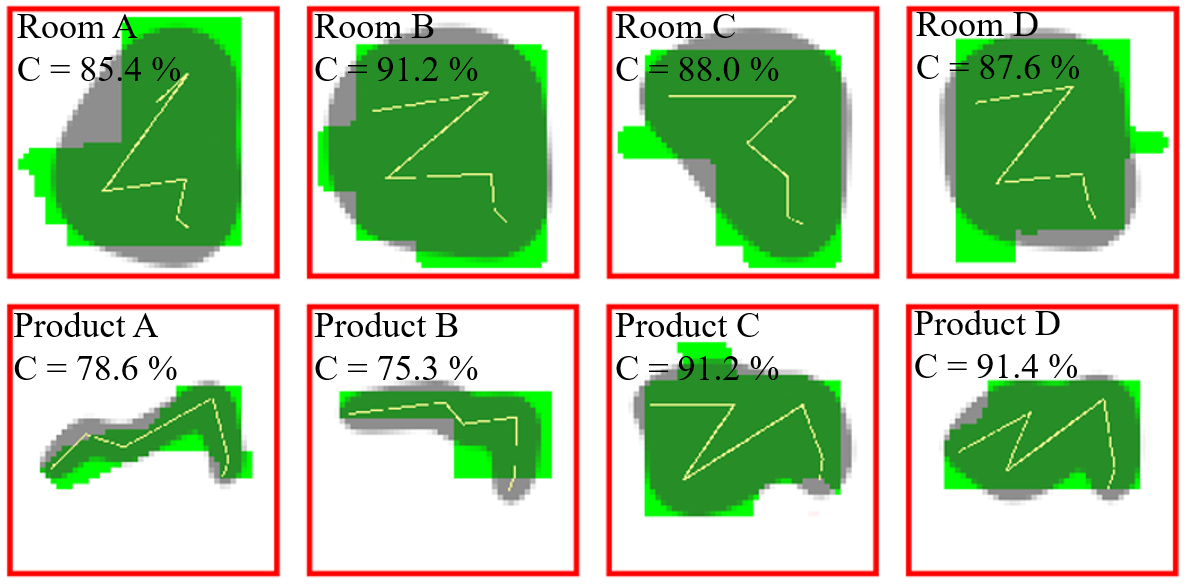}}
	\caption{Output of the trained AI model for unseen target cooling areas. The upper row shows results for living room shapes as used during training. The lower row shows results for actual product geometries. Green: Target area, grey: compressed state of TIM, light-yellow: dispense path. The percentage of covered target area $C$ is indicated for each area.}
	\label{fig:04_results}
\end{figure}

The ANN creates feasible dispense paths, which do not exhibit any voids and maintain a constant dispense feedrate.
Figure~\ref{fig:04_results} shows the compressed state (grey) as overlay onto the target cooling area (green) along with the dispense path (light-yellow).
The percentage of covered target area $C$ is calculated by dividing the sum of covered target area cells by the sum of total target area cells.
The paths can be transferred easily to automated dispensing machines.
The compressed states match the target area well.
All paths were predicted with computation times below one second.

In contrast, \textit{TIMtrace} has computation times of up to one week and achieves following coverage values C: 93.7\,\% for product A, 96.8\,\% for product B, 95.6\,\% for product C, 97.4\,\% for product D.
%Query from SQL id=14190 (A), 14075 (b)
% source: see comments in latex file of timtrace paper

%
%
%
\section{Discussion}\label{sec:discussion}
The AI model is able to learn a design strategy by adapting a certain path shape to a given target area.
This is similar to a human expert, who learns a design strategy from experience.

We achieve good area coverage results -- especially considering that we use simple polygonal chains with a length of only five segments as dispense paths.
We notice drawbacks in result quality for products A and B, which both differ from our used training dataset.
We expect a better generalization from using a training dataset with a higher variety of area shapes.
Non-optimum results could be used as a starting point for an optimization-based approach. 
During inference, our newly proposed approach has strong advantages over ~\textit{TIMtrace} with regard to computation time.

Scholtes et al. design the dispensed pattern (=material distribution after dispensing) as opposed to the dispense path (=path of the dispensing tool).
This involves many degrees of freedom for the design and makes it very difficult to formulate constraints directly on the path.
A simple medial axis transform~\cite{blum_transformation_1967} yields visually similar shapes to the implementation of Scholtes et al. and has low demands regarding implementation efforts and computational resources.
In contrast to Scholtes et al., our proposed setup yields dispense paths that can be easily transferred to manufacturing machines and that natively prevent air entrapments.

%A lack of sufficient data quantity and quality frequently inhibits the successful application of AI based approaches~\cite{}. Our proposed setup can generate process parameters -- in this case the definition of the dispense paths -- without the need for labeled training data.
%Our proposed approach is conceptually transferable to other manufacturing processes if a suitable quality model ANN can be set up.
%
%
%
\section{Conclusion and outlook}
We successfully train an AI model to predict TIM dispense paths without needing labeled training data.
The resulting AI model can generate designs for dispense paths on unseen target cooling areas.
Inference can be carried out in real-time.
The generated paths are feasible and fully compatible with the manufacturing constraints of automated dispensing machines in series manufacturing.

We aim to study the transferability of this approach to other manufacturing processes, since it essentially enables an inline optimization of process parameters.
\section*{Acknowledgments}
Patents~\cite{baeuerle_waerme_2023, baeuerle_heat_2023, baeuerle_prozess_2023, baeuerle_process_2023} of Baeuerle et al. apply.
%
%
%
%\section*{Author statement}
%%
%We describe the individual contributions of Simon Baeuerle (SB), Ian Franson Mendonca (IM), Kristof Van Laerhoven (KL), Andreas Steimer (AS) and Ralf Mikut (RM) using CRediT~\cite{brand_beyond_2015}: \textit{Writing - Original Draft}: SB; \textit{Writing - Review \& Editing}: IM, KL, AS, RM; \textit{Conceptualization}: SB, AS, RM; \textit{Investigation}: SB, IM; \textit{Methodology}: SB, AS; \textit{Software}: SB, IM; \textit{Supervision}: KL, AS, RM; \textit{Project Administration}: SB, RM; \textit{Funding Acquisition}: SB, RM.
%%

\bibliographystyle{IEEEtran}
%\bibliography{literature_raw}

\end{document}